\documentclass[runningheads]{llncs}

\usepackage{url}

\usepackage[T1]{fontenc}

\usepackage{amsmath}
\usepackage{amssymb}
\usepackage{booktabs}
\usepackage{graphicx}
\usepackage[hidelinks]{hyperref}
\usepackage{multirow}
\usepackage{physics}
\usepackage{subcaption}
\usepackage{tabularx}
\usepackage{xcolor}

\definecolor{fireenginered}{rgb}{0.81, 0.09, 0.13}

\newcommand{\Arxiv}{}

\begin{document}
\mainmatter
\title{On Robust Reinforcement Learning with Lipschitz-Bounded Policy Networks}
\titlerunning{Robust RL with Lipschitz policy networks}
\author{Nicholas H. Barbara \and Ruigang Wang \and Ian R. Manchester}
%
%
\authorrunning{N. H. Barbara et al.}
\tocauthor{Nicholas H. Barbara, Ruigang Wang, and Ian R. Manchester}
\institute{Australian Centre for Robotics, School of Aerospace, Mechanical and Mechatronic Engineering, The University of Sydney, Sydney, NSW 2006, Australia.\\
\email{\{nicholas.barbara, ruigang.wang, ian.manchester\}@sydney.edu.au}}

\maketitle

%
%

\begin{abstract}
This paper presents a study of robust policy networks in deep reinforcement learning. We investigate the benefits of policy parameterizations that naturally satisfy constraints on their Lipschitz bound, analyzing their empirical performance and robustness on two representative problems: pendulum swing-up and Atari Pong. We illustrate that policy networks with smaller Lipschitz bounds are more robust to disturbances, random noise, and targeted adversarial attacks than unconstrained policies composed of vanilla multi-layer perceptrons or convolutional neural networks. However, the structure of the Lipschitz layer is important. We find that the widely-used method of spectral normalization is too conservative and severely impacts clean performance, whereas more expressive Lipschitz layers such as the recently-proposed Sandwich layer can achieve improved robustness without sacrificing clean performance.
\end{abstract}

%
%
\section{Introduction} \label{sec:intro}

Deep reinforcement learning (deep RL) has been the driving force behind many recent successes in learning-based control, including in discrete game-like problems \cite{Mnih++2015}, robotic manipulation \cite{Kalashnikov++2018}, and locomotion \cite{Rudin++2021}. However, the applicability of deep RL to performance- and safety-critical systems is currently limited by questions of its robustness \cite{Huang++2017}. It is well known that neural networks can be highly sensitive to small input perturbations \cite{Szegedy++2013}, making policy networks learned via deep RL potentially unrobust to disturbances, noise, and targeted adversarial attacks. Despite sharing similar sensitivity issues to neural classifiers, for which many robust neural networks have recently been developed \cite{Bernd+Lampert2022,Trockman+Kolter2021,Wang+Manchester2023}, the use of robust policy architectures in deep RL has not been widely studied.

Most common approaches to improving policy robustness in deep RL are based on adversarial training \cite{Pattanaik++2017}, where adversarial attacks on a policy's inputs are optimized during training to encourage the network to perform well under perturbations. While this works well in applications where the structure of the perturbations is always similar to those seen during training, adversarial training only certifies a lower bound on a policy's robustness. It is therefore possible to find new, out-of-sample attacks that cause the policy to fail \cite{Russo+Proutiere2021}. An alternative strategy is to learn control policies with a certified \textit{upper} bound on their sensitivity to perturbations using methods like randomized smoothing \cite{Kumar++2021,Wu++2021} and loss-function regularization \cite{Oikarinen++2021,Nie++2024}. These methods bound the sensitivity of a learned policy during the training process. It is interesting to ask whether, via careful choice of a policy's architecture and parameterisation, we can directly bound the sensitivity of a policy independently of how it is trained.

\begin{figure}[!t]
    \centering
    \includegraphics[trim={6.5cm 12.1cm 8cm 4.1cm},clip,width=0.65\textwidth]{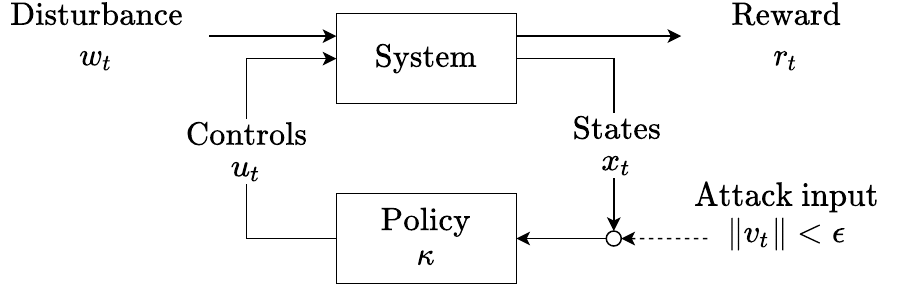}
    \caption{Reinforcement learning and adversarial attacks.}
    \label{fig:rl-summary}
    \vspace{-1mm}
\end{figure}

One promising approach is to use neural network policies that are certifiably robust to perturbations \textit{by construction}. This can be achieved by constraining the Lipschitz bound of the network. A neural network $f:\mathbb{R}^n\rightarrow \mathbb{R}^m$ is said to have an $\ell_2$ Lipschitz bound of $\gamma$ if
\begin{equation} \label{eqn:lipschitz}
    \|f(x_1) - f(x_2)\|_2 \le \gamma \|x_1 - x_2\|_2, \quad \forall x_1,x_2\in \mathbb{R}^n
\end{equation}
The true $\ell_2$ Lipschitz constant, denoted $\mathrm{Lip}(f)$, is the smallest $\gamma$ satisfying \eqref{eqn:lipschitz}. Lipschitz-bounded networks are ``smoother'' and less sensitive to input perturbations than unconstrained networks since small variations in their inputs only induce small variations in their outputs. Despite the wealth of recent work on constructing Lipschitz-bounded deep networks \cite{Pauli++2022,Trockman+Kolter2021,Wang+Manchester2023}, to the best of our knowledge, the only method tested in deep RL is spectral normalization \cite{Bjorck++2021,Takase++2022} which is known to give conservative bounds on the true Lipschitz constant. 

This paper therefore investigates the following questions:
\begin{enumerate}
    \item Can Lipschitz-bounded deep networks improve the empirically-observed robustness of control policies in deep RL? 
    \item If so, does the policy network architecture matter? That is, do more sophisticated policy parameterizations with less-conservative Lipschitz bounds give finer control over the performance-robustness trade-off?
\end{enumerate}
We provide an initial empirical study of Lipschitz-bounded policy networks in deep RL on two representative problems: pendulum swing-up, a classical benchmark problem in control and RL; and Atari Pong, a simple proxy for vision-based autonomous decision-making tasks. To the best of our knowledge, our comparison of Lipschitz-bounded policy architectures in deep RL is the first of its kind.

%
%
\section{Background and Prior Work} \label{sec:background}

In this section, we first introduce the concepts of deep reinforcement learning and adversarial attacks. We then provide an overview of recent developments in robust neural networks with certifiable Lipschitz bounds.

\subsection{Deep Reinforcement Learning} \label{sec:background-rl}
\vspace{-1mm}

We introduce RL from a control-theoretic perspective (Fig.~\ref{fig:rl-summary}).
Consider a discrete-time nonlinear dynamical system
\begin{align} \label{eqn:dynamics}
    x_{t+1} &= f(x_t, u_t, w_t)
\end{align}
with state vector $x_t\in\mathbb{R}^n$, controlled inputs $u_t\in\mathbb{R}^m$, and disturbances $w_t\in\mathbb{R}^p$. The task is to learn feedback control policies of the form $u_t = \kappa(x_t; \theta)$ parameterized by $\theta\in\mathbb{R}^q$ which (locally and approximately) solve 
\begin{equation} \label{eqn:rl-problem}
\begin{aligned}
    &\max_{\theta} \ R := \mathop{\mathbb{E}}_{x_0, w} \left[ \sum_{t=0}^{\infty} \rho^t r_t(x_t, u_t)\right] \,
    \text{s.t.} \, \ x_{t+1} = f(x_t, u_t, w_t), \ u_t = \kappa(x_t; \theta),
\end{aligned}
\end{equation}
where $r_t$ is the instantaneous reward, $\rho\in(0,1)$ is a ``discount factor,'' $R$ is the total expected reward, and the expectation $\mathbb{E}[\cdot]$ is taken over some known distributions of disturbances $w_t$ and initial conditions $x_0$. In deep RL, the controller $\kappa$ is a deep neural network (DNN) parameterized by $\theta$. In this paper we take Proximal Policy Optimization (PPO) \cite{Schulman++2017} as our training method due to its simplicity, speed, and performance.

\subsection{Adversarial Attacks for RL} \label{sec:background-attacks}
\vspace{-1mm}

Given a learned policy $\kappa(\cdot\,;\theta)$, an adversarial attack is an input sequence $v = (v_0, v_1, \ldots)$ with some restricted ``attack size'' $\epsilon > 0$ that is designed to reduce the expected cumulative reward of the policy as much as possible. While attacks can be designed to change the closed-loop behavior in many ways, the most common structure is an additive perturbation to the policy input (i.e., perturbations to the state measurements, Fig.~\ref{fig:rl-summary}) \cite{Huang++2017,Pattanaik++2017}. The problem can be formulated as 
\begin{equation} \label{eqn:optimal-attack}
\begin{aligned}
    \min_{v}\ & R \quad 
    \text{s.t.}\quad x_{t+1} = f(x_t, \kappa(x_t + v_t; \theta), w_t),\quad \norm{v_t} \le \epsilon,\; \forall t,
\end{aligned}
\end{equation}
where $\norm{\cdot}$ can be any $p$-norm with $1\leq p\leq \infty$. If the reward function $r_t$ and dynamic model $f$ are both differentiable and known to the attacker, \eqref{eqn:optimal-attack} can be solved by gradient descent methods. If part of the dynamical system is unknown or not differentiable, solving (4) is difficult or impossible and it is common instead to solve the simplified problem
\begin{equation} \label{eqn:attack-pgd}
    \max_{v_t}\quad \|\kappa(x_t+v_t;\theta)-\kappa(x_t; \theta)\| \quad \mathrm{s.t.}\quad \|v_t\|\leq \epsilon.
\end{equation}
That is, find an admissible attack $v_t$ which leads to large policy output perturbation at each time step, regardless of its effect on the final reward \cite{Huang++2017}. Problem~\eqref{eqn:attack-pgd} can be solved by attack methods like Projected Gradient Descent (PGD) \cite{Madry++2018}.

\subsection{Lipschitz-Bounded Deep Networks} \label{sec:background-lipnets}
\vspace{-1mm}

The adversarial attack problem in \eqref{eqn:attack-pgd} is exactly the calculation of the (local) Lipschitz constant of the policy network $\kappa$. We can therefore control the effect of adversarial attacks everywhere in state space by bounding the global Lipschitz constant $\mathrm{Lip}(\kappa) \le \gamma$. Policy networks with smaller $\gamma$ are then ``smoother'' and are likely to be more robust to small attacks than those with a large $\gamma$.

In deep RL, the policy network $\kappa$ is often parameterized by a multi-layer perceptron (MLP) or a convolutional neural network (CNN) of the form of 
\begin{equation}\label{eq:mlp}
    \kappa=g_L \circ \sigma \circ g_{L-1}\circ \cdots \circ \sigma \circ g_1
\end{equation}
where $\sigma$ is a fixed monotone and 1-Lipschitz scalar activation (e.g., ReLU, tanh, sigmoid, etc.) and $g_k$ is a linear layer $g_k(x)=W_kx + b_k$ with $W_k,b_k$ as the weights and biases, respectively. The Lipschitz constraint for \eqref{eq:mlp} is then
\begin{equation}
    \mathrm{Lip}(\kappa)=\sup_{J_2,\ldots,J_{L}} \| W_{L} J_{L} W_{L-1}\cdots J_2 W_1\|_2\leq \gamma
\end{equation}
where $J_k$ is a diagonal matrix with $0\leq J_{k,ii}\leq 1$. Since it is NP-hard to compute the exact Lipschitz constant \cite{virmaux2018lipschitz}, a practical approach is to find an upper bound $\overline{\gamma}$ for $\mathrm{Lip}(\kappa)$ and then impose the constraint $\overline{\gamma}=\gamma$. There are many bound estimation methods resulting in different constructions of Lipschitz networks. A metric to measure the expressive power of Lipschitz networks is \emph{tightness}, which is defined as $\underline{\gamma}/\gamma$ \cite{Wang+Manchester2023} with $\underline{\gamma}$ as the empirical lower Lipschitz bound:
\begin{equation} \label{eqn:lipschitz-lower}
    \underline{\gamma}:=\max_{x\in\mathbb{X},\|v\|_2\leq \epsilon} \|\kappa(x+v)-\kappa(x)\|_2/\|v\|_2
\end{equation}
where $v\in\mathbb{R}^n$ and $\mathbb{X} \subset \mathbb{R}^n$ is some compact region. Note that $\mathrm{Lip}(\kappa)\in[\underline{\gamma},\gamma]$.

A simple Lipschitz bound estimation is the layer-wise spectral norm bound $ \overline{\gamma}_{s}=\prod_{k=1}^{L}\|W_k\|_2$. This approach uses the fact that common activation functions $\sigma$ are 1-Lipschitz. Thus, we can construct Lipschitz policy networks via scaling factors and 1-Lipschitz linear layers, where three representative examples are given as follows.
 
\begin{itemize}
\item[$\bullet$] \textbf{(SN)} The spectral normalization (SN) layer \cite{miyato2018spectral} is a linear layer with weight $W=(1/\rho) A$ where $A$ is a learnable weight, and $\rho$ is its maximal singular value (i.e., $\|W\|_2=1$). Lipschitz networks composed of SN layers often have quite loose Lipschitz bounds such that $\underline{\gamma}/\gamma$ is very small.

\item[$\bullet$] \textbf{(AOL)} The almost orthogonal Lipschitz (AOL) layer \cite{Bernd+Lampert2022} is a linear layer with weight $ W=A D $ where $A$ is a learnable matrix and $D$ is a diagonal matrix with $D_{ii}=\sqrt{\sum_{j}|A^\top A|_{ij}}.$ The experimental results in \cite{Bernd+Lampert2022} show that the learned weight $W$ tends to have singular values close to 1, which helps to improve the model tightness.

\item[$\bullet$] \textbf{(Cayley)} To obtain an even tighter Lipschitz bound, an orthogonal layer was proposed in \cite{Trockman+Kolter2021} by leveraging the Cayley transform. Specifically, given a free learnable weight $P\in \mathbb{R}^{n\times n}$, one first obtains a skew symmetric matrix $A=P-P^\top$ and then constructs the weight matrix by $W=(I-A)(I+A)^{-1}.$ It is easy to verify that $W^\top W=I$ and so all singular values of $W$ are 1.
\end{itemize}

A tighter and more expressive bound estimation method was recently proposed in \cite{Fazlyab++2019}, which explores both the monotonicity and Lipschitz properties of the activation $\sigma$ by leveraging the integral quadratic constraint (IQC) framework \cite{megretski1997system}. Direct (i.e. unconstrained) parameterizations based on IQC were proposed in \cite{Revay++2020} for deep equilibrium networks, in \cite{araujo2023unified} for residual networks, and in \cite{Wang+Manchester2023} for deep MLPs and CNNs. In particular, \cite{Wang+Manchester2023} shows that the IQC-based approach can lead to a 1-Lipschitz nonlinear layer as follows.
\begin{itemize}
    \item[$\bullet$] \textbf{(Sandwich)} The Sandwich layer is a 1-Lipschitz nonlinear layer of the form
    \begin{equation}\label{eq:sandwich}
        g(x)= \sqrt{2}A^\top \Psi \sigma\bigl(\sqrt{2}\Psi^{-1} Bx+b\bigr)
    \end{equation}
    where $Q=\begin{bmatrix}
        A & B
    \end{bmatrix}$ is a semi-orthogonal matrix ($QQ^\top=I$) parameterized by the Cayley transformation, and $\Psi$ is a positive diagonal matrix parameterized via an exponential mapping. Based on the above 1-Lipschitz layer, we can construct the $\gamma$-Lipschitz policy network $\kappa$ via 
    \begin{equation}\label{eq:lbdn}
        \kappa(x) = \sqrt{\gamma}g_{L}\circ g_{L-1}\circ \cdots \circ g_1(\sqrt{\gamma}x).
    \end{equation}
    Note that \eqref{eq:lbdn} can be transformed back to the MLP form \eqref{eq:mlp}. 
\end{itemize}

The Sandwich layer \eqref{eq:sandwich} is an ``expressive'' layer architecture in that it contains all 1-Lipschitz linear layers as a special case, and it allows the spectral norm bound of each layer (and their product) to be greater than one \cite{Wang+Manchester2023}.
Experimental results in \cite{Wang+Manchester2023} on image datasets show that \eqref{eq:lbdn} can achieve better performance than 1-Lipschitz linear layers such as AOL and Cayley. It is natural to ask whether similar results hold true for deep RL.

%
%
\section{Experimental Setup} \label{sec:setup-methods}

We study two classic RL problems --- pendulum swing-up and Atari Pong --- to investigate the research questions outlined in Section~\ref{sec:intro}. Our code and training details are available on GitHub\footnote{\url{https://github.com/nic-barbara/Lipschitz-RL-MJX}}$^\text{,}$\footnote{\url{https://github.com/nic-barbara/Lipschitz-RL-Atari}}%
\ifdefined\Arxiv
$\ $and in Appendix~\ref{app:training-details}, respectively.
\else
.
\fi


\textbf{Pendulum Swing-up} aims to swing a physical pendulum to its upright equilibrium based on the quadratic reward $r_t=-(\alpha_t^2 + 0.1\dot{\alpha}_t^2 + 0.001u_t^2)$, where $\alpha_t$ is the pendulum angle (wrapped to $[-\pi,\pi]$) and $u_t$ is the pendulum torque.  The optimal policy is well-known to have sharp decision boundaries which make it susceptible to chattering and instability under small measurement perturbations, delays, or uncertainty. We trained MLP policies \eqref{eq:mlp} without any bounds on the network Lipschitz constant, and Lipschitz-bounded policies using the Sandwich parameterization \eqref{eq:sandwich}. We investigated the robustness of each policy to two sources of perturbations: a) sample delays; and b) adversarial attacks with constrained $\ell_2$ norm. Adversarial attacks were computed by solving \eqref{eqn:optimal-attack} over a sequence of four 50-sample windows with gradient descent.


\textbf{Atari Pong} is a video game in which two players each control a paddle that can move up and down and try to deflect a puck into their opponent's goal (Fig.~\ref{fig:pong-gameplay-attacks}). The reward is the net game score, with a maximum score of 21 goals to nil. An automated ``computer'' player controls the left paddle and the RL policy controls the right paddle. It is a commonly-studied benchmark in deep RL \cite{Mnih++2015,Russo+Proutiere2021}. The game can be written as an RL problem \eqref{eqn:rl-problem} where the states $x_t$ are grayscale gameplay images and the control actions $u_t$ are discrete paddle movements. We trained a classic, unconstrained CNN model \eqref{eq:mlp} and four different Lipschitz-bounded policy architectures (SN, AOL, Sandwich, Cayley) with various Lipschitz bounds. We compared the robustness of each policy network to: a) uniform random noise; b) PGD attacks \eqref{eqn:attack-pgd} with constrained $\ell_2$ norm; and c) PGD attacks \eqref{eqn:attack-pgd} with constrained $\ell_\infty$ norm.

%
%
\section{Results and Discussion} \label{sec:results}

We first study the advantages of Lipschitz-bounded policy networks in terms of robustness to perturbations and adversarial attacks, using pendulum swing-up as an illustrative example (Sec.~\ref{sec:pendulum-results}). We extend this study to vision-based feedback control in Pong (Sec.~\ref{sec:pong-results}) and compare the benefits of different Lipschitz-bounded policy architectures. In Figures~\ref{fig:pendulum-attacks}, \ref{fig:pong-attacks}, and Table~\ref{tab:pong-robustness}, Lipschitz lower bounds $\underline\gamma$ were computed for each policy by performing gradient ascent on \eqref{eqn:lipschitz-lower}.

\subsection{Illustrative Example --- Pendulum Swing-up} \label{sec:pendulum-results}

\begin{figure}[ht]
    \centering
    \begin{subfigure}[b]{0.445\textwidth}
        \centering
        \includegraphics[trim={10cm 4cm 10cm 3.5cm},clip,width=\textwidth]{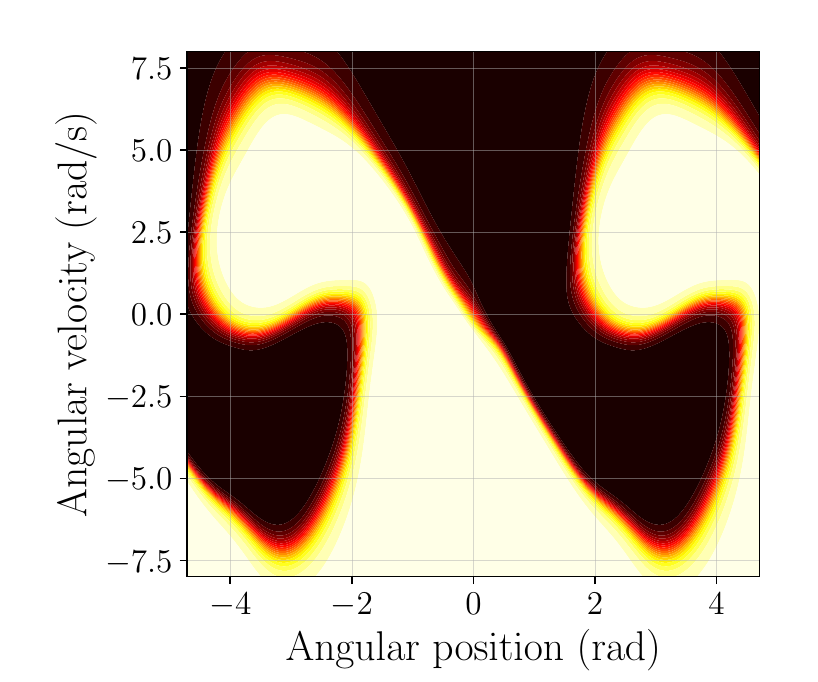}
        \caption{Unconstrained policy landscape.}
        \label{fig:pendulum-contours-mlp}
    \end{subfigure}
    \begin{subfigure}[b]{0.54\textwidth}
        \centering
        \includegraphics[trim={8cm 4cm 9cm 3.5cm},clip,width=\textwidth]{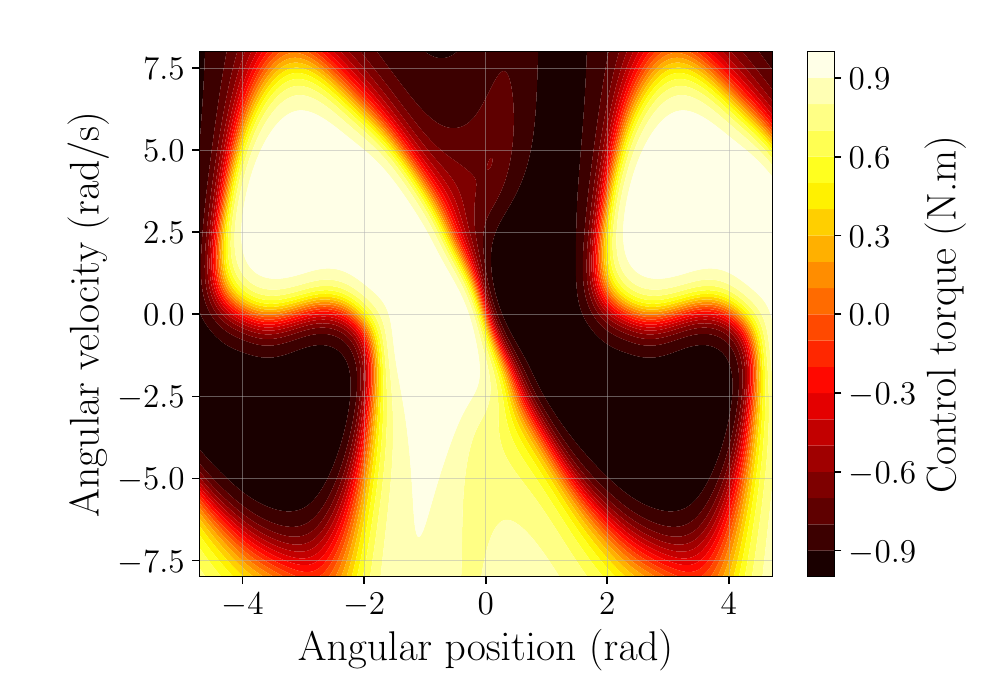}
        \caption{Lipschitz-bounded  policy landscape.}
        \label{fig:pendulum-contours-lbdn}
    \end{subfigure}
    \begin{subfigure}[b]{0.445\textwidth}
        \centering
        \includegraphics[trim={10cm 4cm 10cm 3.5cm},clip,width=\textwidth]{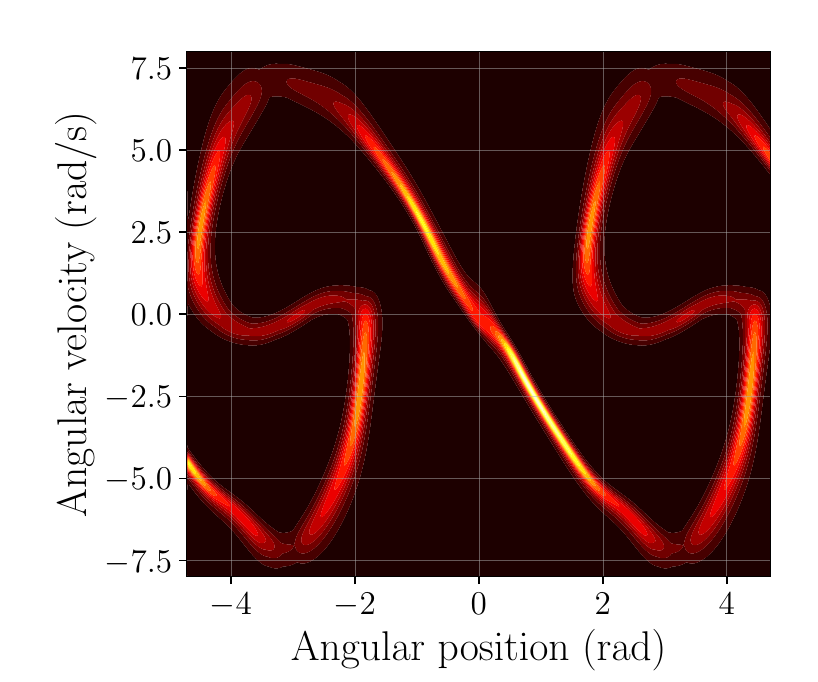}
        \caption{Local Lipschitz (unconstrained).}
        \label{fig:pendulum-contours-mlp-lip}
    \end{subfigure}
    \begin{subfigure}[b]{0.54\textwidth}
        \centering
        \includegraphics[trim={8cm 4cm 9cm 3.5cm},clip,width=\textwidth]{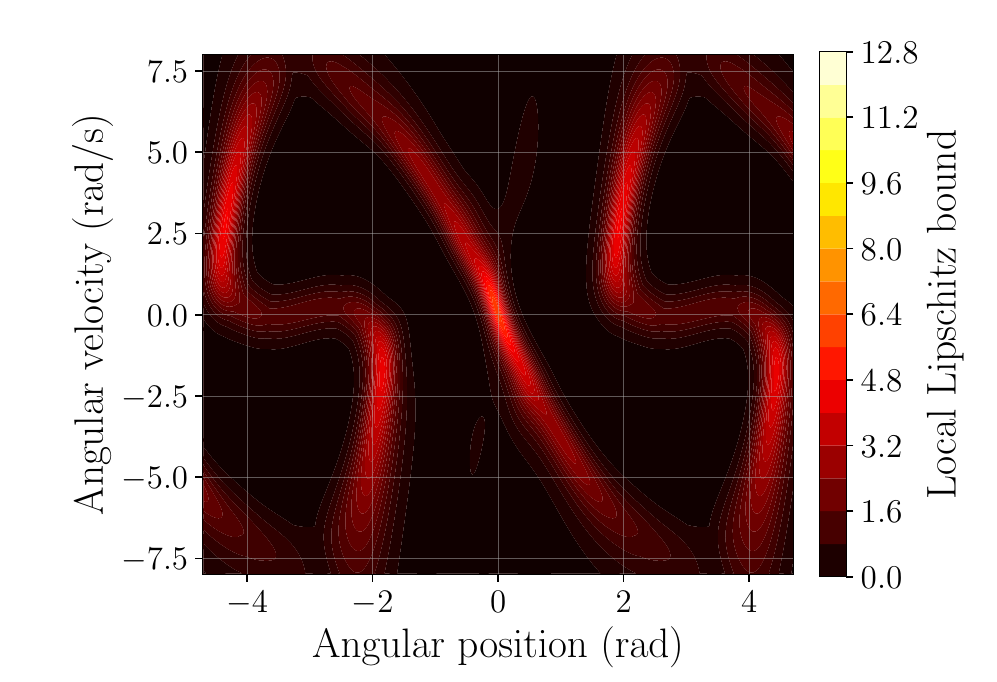}
        \caption{Local Lipschitz (Lipschitz-bounded).}
        \label{fig:pendulum-contours-lbdn-lip}
    \end{subfigure}
    \caption{Contours of control actions (a,b) and local Lipschitz bounds (c,d) for an unconstrained (MLP) and a Lipschitz-bounded (Sandwich, $\gamma = 4$) policy show how Lipschitz bounds control a policy's smoothness. 
    }
    \label{fig:pendulum-contours}
\end{figure}

Let us first consider the effect of small Lipschitz bounds on unperturbed policy networks. Figure~\ref{fig:pendulum-contours} compares the policy landscape and (empirically-estimated) local Lipschitz constant in phase space for an unconstrained MLP policy and a Lipschitz-bounded policy composed of Sandwich layers with $\gamma = 4$. The unconstrained policy has sharp decision boundaries, either side of which is flips the sign of the control torque (limited to $\pm1$\,N.m). While these sharp changes are optimal in the unperturbed case, it is clear that any small uncertainty in the pendulum's position or velocity will cause the network to apply a drastically different control action, potentially driving the system to instability. In contrast, imposing a bound on the Lipschitz constant with Sandwich layers visibly smooths the decision boundaries with negligible penalty to the final test reward ($-153$ for unconstrained and $-157$ for Lipschitz-bounded).

\begin{figure}[!t]
     \centering
     \begin{subfigure}[b]{0.495\textwidth}
         \centering
         \includegraphics[trim={0cm 0.7cm 0cm 0.5cm},clip,width=\textwidth]{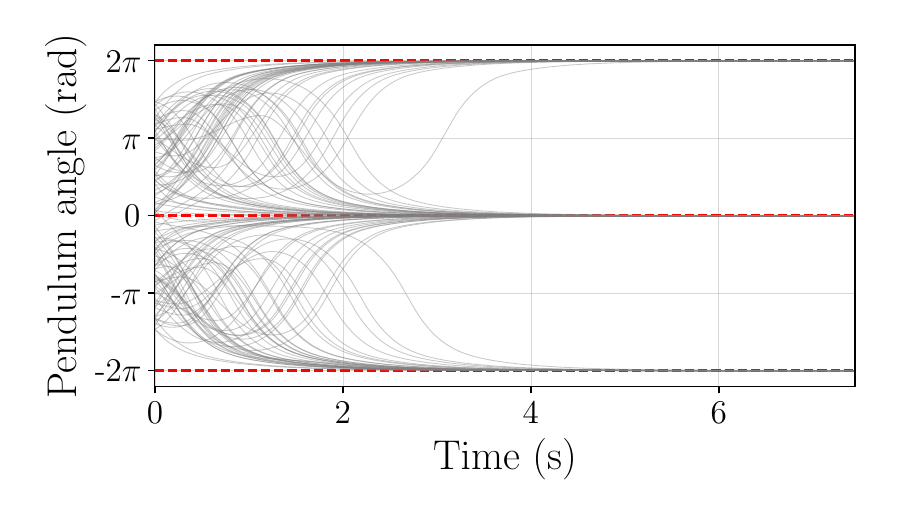}
         \caption{Unconstrained.}
         \label{fig:pendulum-trajectories-mlp}
     \end{subfigure}
     \begin{subfigure}[b]{0.495\textwidth}
         \centering
         \includegraphics[trim={0cm 0.7cm 0cm 0.5cm},clip,width=\textwidth]{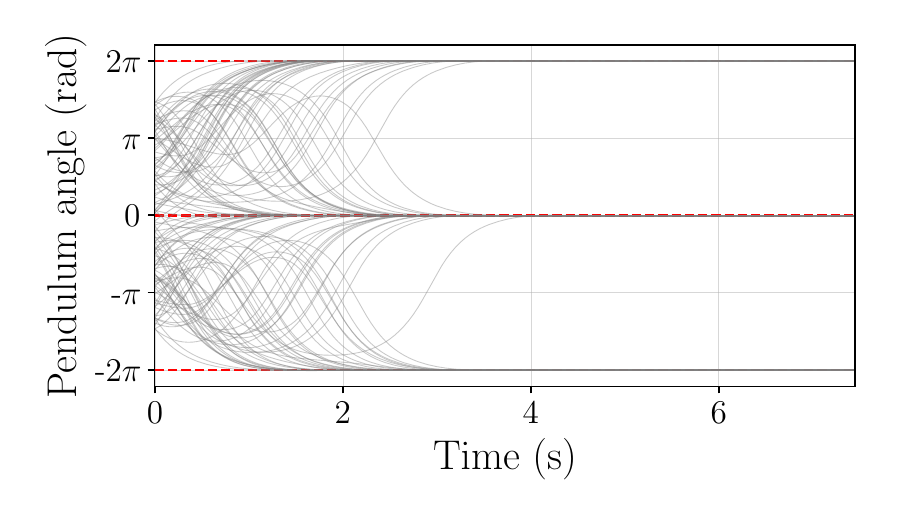}
         \caption{Lipschitz-bounded.}
         \label{fig:pendulum-trajectories-lbdn}
     \end{subfigure}
          \begin{subfigure}[b]{0.495\textwidth}
         \centering
         \includegraphics[trim={0cm 0.7cm 0cm 0.5cm},clip,width=\textwidth]{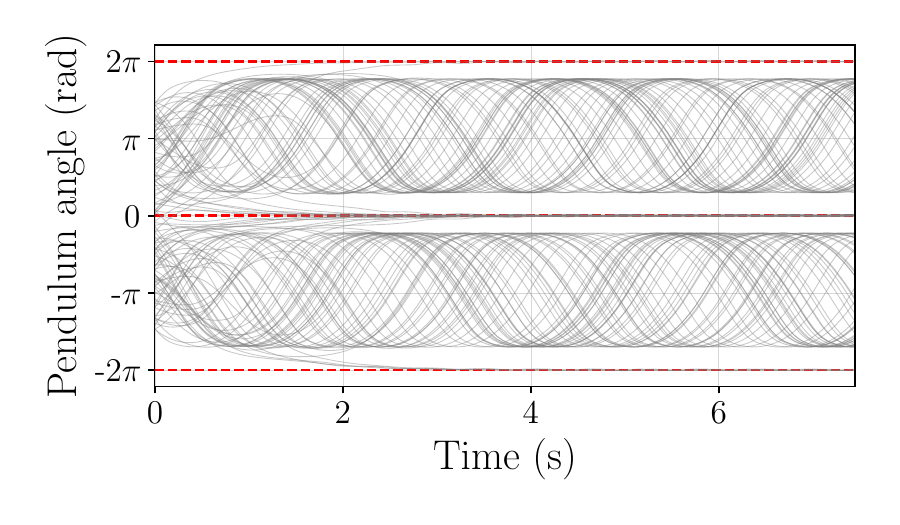}
         \caption{Unconstrained (0.1\,s delay).}
         \label{fig:pendulum-trajectories-mlp-delay}
     \end{subfigure}
     \begin{subfigure}[b]{0.495\textwidth}
         \centering
         \includegraphics[trim={0cm 0.7cm 0cm 0.5cm},clip,width=\textwidth]{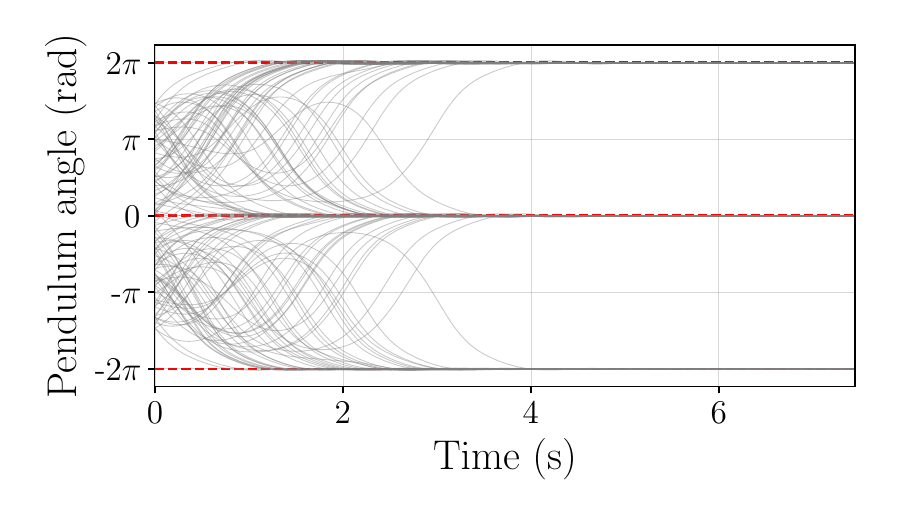}
         \caption{Lipschitz-bounded (0.1\,s delay).}
         \label{fig:pendulum-trajectories-lbdn-delay}
     \end{subfigure}
     \begin{subfigure}[b]{0.495\textwidth}
         \centering
         \includegraphics[trim={0cm 0.7cm 0cm 0.5cm},clip,width=\textwidth]{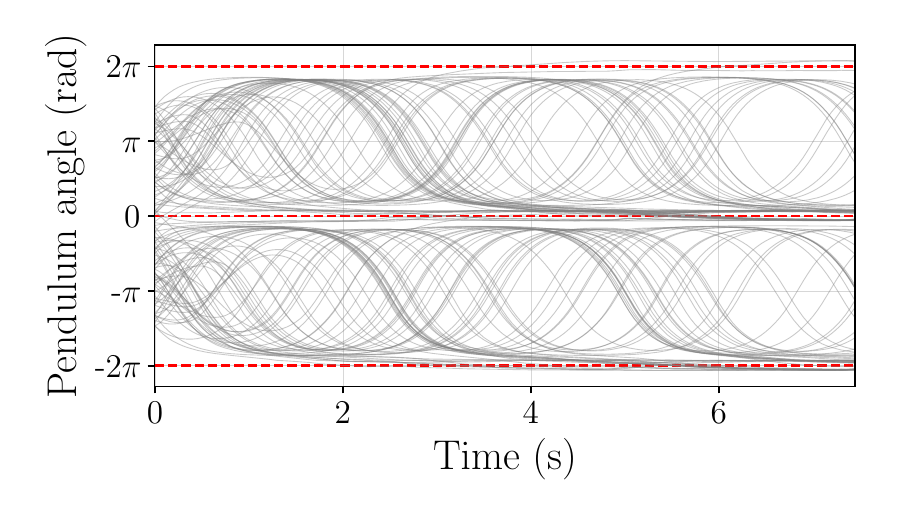}
         \caption{Unconstrained ($\ell_2$ attack, $\epsilon = 0.11$).}
         \label{fig:pendulum-trajectories-mlp-attacked}
     \end{subfigure}
     \begin{subfigure}[b]{0.495\textwidth}
         \centering
         \includegraphics[trim={0cm 0.7cm 0cm 0.5cm},clip,width=\textwidth]{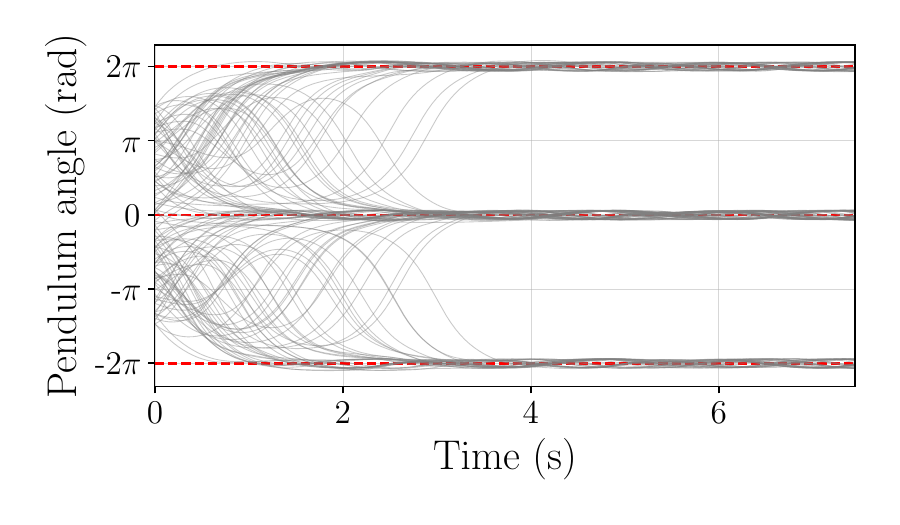}
         \caption{Lipschitz-bounded ($\ell_2$ attack, $\epsilon = 0.11$).}
         \label{fig:pendulum-trajectories-lbdn-attacked}
     \end{subfigure}
    \caption{Pendulum trajectories generated by unconstrained (MLP) and Lipschitz-bounded (Sandwich, $\gamma = 4$) policies in nominal operation (a,b), with sample delays (c,d), and with $\ell_2$ adversarial attacks (e,f). Red lines indicate the target.}
    \label{fig:pendulum-trajectories}
\end{figure}

Figure~\ref{fig:pendulum-trajectories} illustrates the effect of this smoothing on each policy's robustness to sample delays and $\ell_2$-optimal adversarial attacks. The simulated trajectories in Figures~\ref{fig:pendulum-trajectories-mlp} and \ref{fig:pendulum-trajectories-lbdn} indicate that both policy networks perform similarly in nominal operation. However, when introducing a small sample delay (2 time samples or 0.1\,s, Figs.~\ref{fig:pendulum-trajectories-mlp-delay} \& \ref{fig:pendulum-trajectories-lbdn-delay}) or a small adversarial attack ($\epsilon = 0.11$, Figs.~\ref{fig:pendulum-trajectories-mlp-attacked} \& \ref{fig:pendulum-trajectories-lbdn-attacked}), the unconstrained policy is unable to hold the pendulum stable and upright, whereas the Lipschitz-bounded policy is successful and only exhibits minor oscillations about the equilibrium under adversarial attacks.

\begin{figure}[ht]
     \centering
     \begin{subfigure}[b]{0.495\textwidth}
         \centering
         \includegraphics[trim={0cm 4cm 0cm 0.5cm},clip,width=\textwidth]{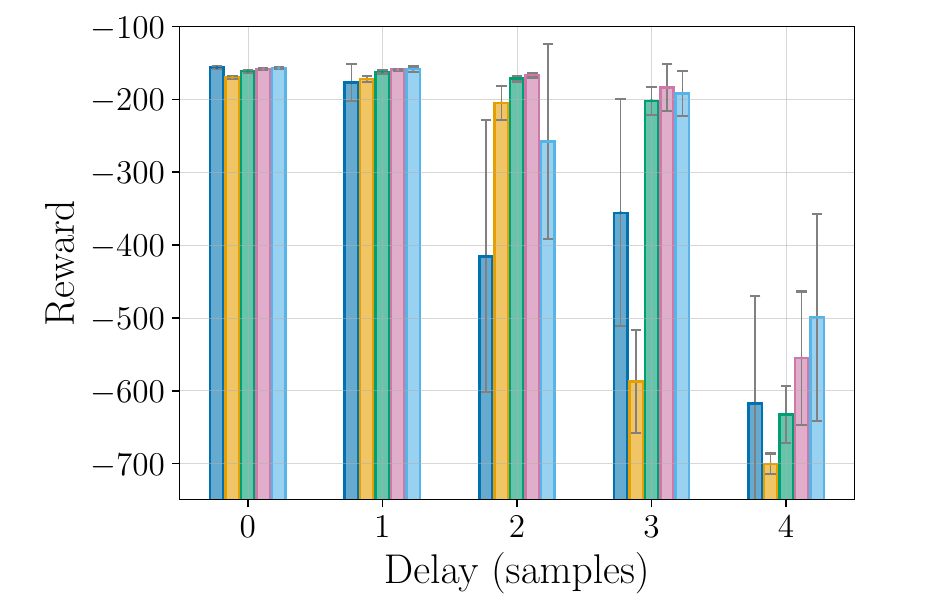}
         \caption{Rewards under sample delay.}
         \label{fig:pendulum-delay-rewards}
     \end{subfigure}
     \begin{subfigure}[b]{0.495\textwidth}
         \centering
         \includegraphics[trim={0cm 4cm 0cm 0.5cm},clip,width=\textwidth]{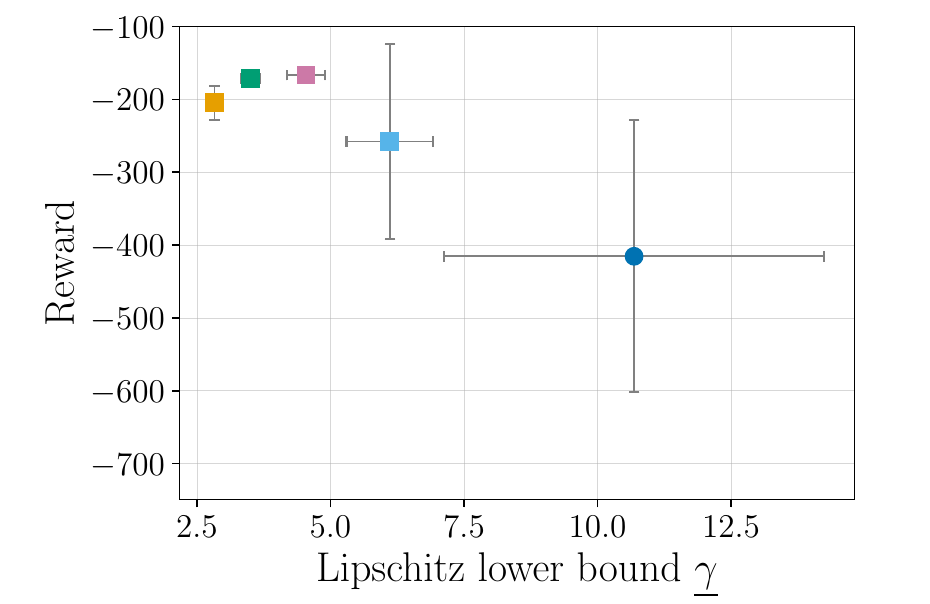}
         \caption{Rewards at 2-step delay (0.1\,s).}
         \label{fig:pendulum-delay-lip}
     \end{subfigure}
     \begin{subfigure}[b]{0.495\textwidth}
         \centering
         \includegraphics[trim={0cm 1.5cm 0cm 0cm},clip,width=\textwidth]{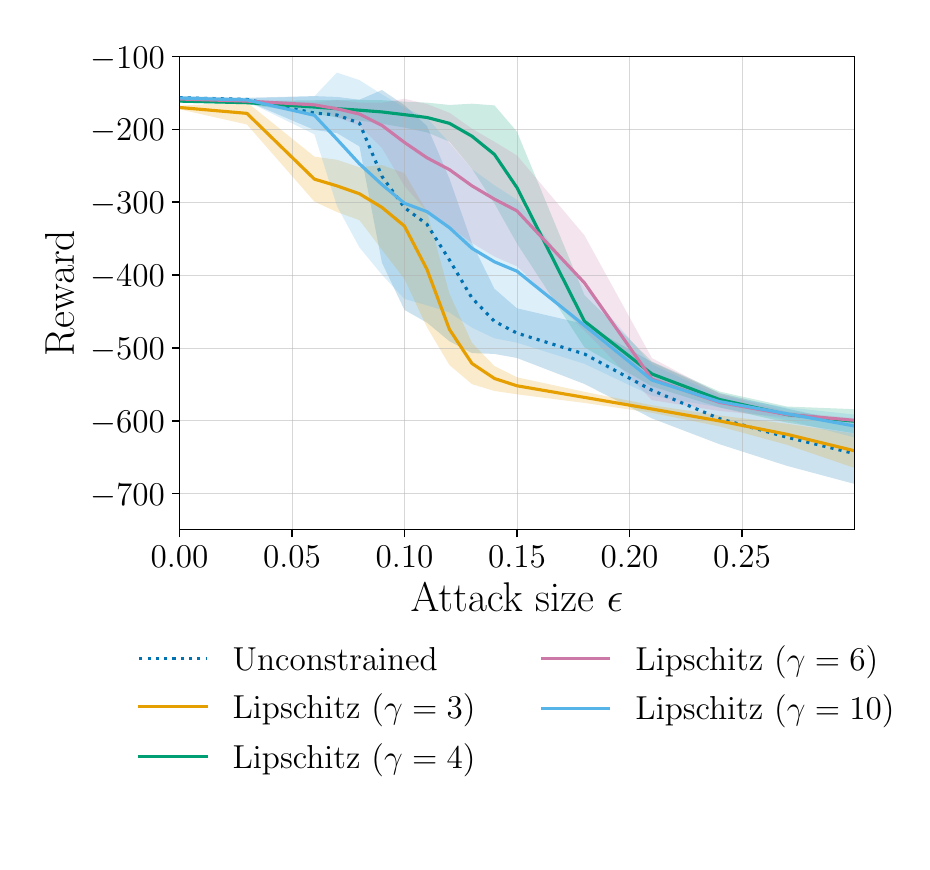}
         \caption{Reward under $\ell_2$-optimal attacks.}
         \label{fig:pendulum-attacks-rewards}
     \end{subfigure}
     \begin{subfigure}[b]{0.495\textwidth}
         \centering
         \includegraphics[trim={0cm 1.5cm 0cm 0cm},clip,width=\textwidth]{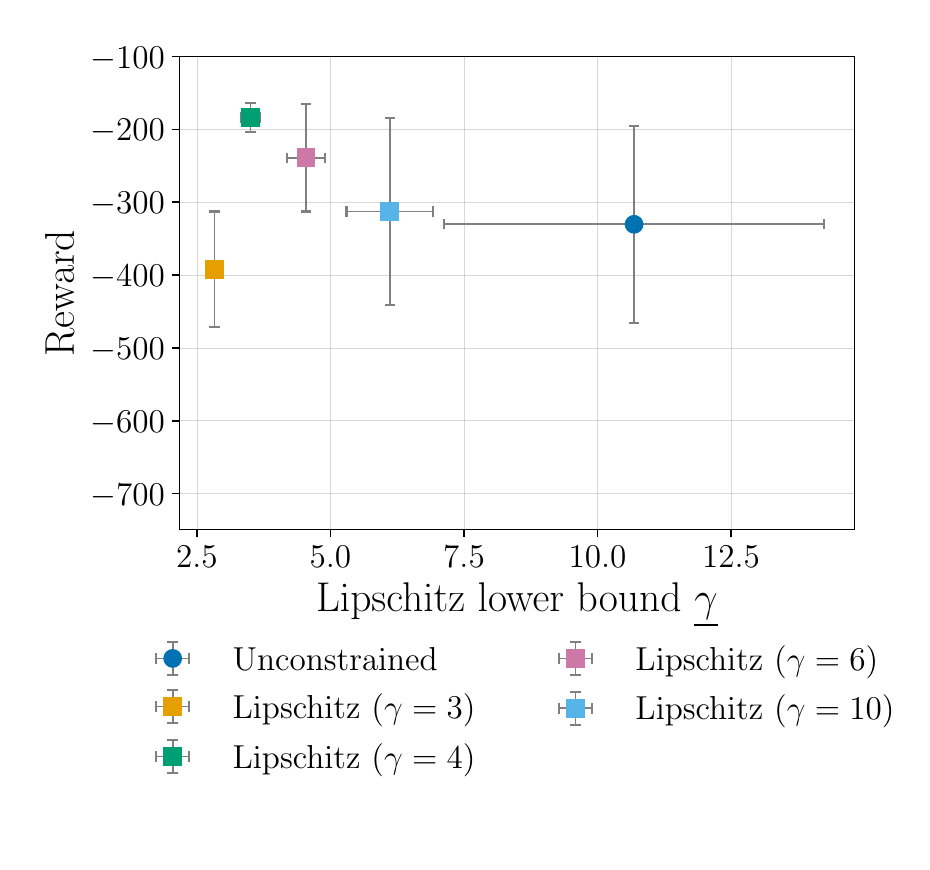}
         \caption{Rewards at $\epsilon = 0.11$.}
         \label{fig:pendulum-attacks-lip}
     \end{subfigure}
    \caption{Robust performance of unconstrained (MLP) and Lipschitz-bounded (Sandwich) policies on pendulum swing-up under sample delays and $\ell_2$-optimal adversarial attacks. Panels (b,d) show cross-sections of (a,c) as a function of each model's empirically-estimated Lipschitz lower bound. Bands and error bars show one standard deviation over 10 random model initializations.}
    \label{fig:pendulum-attacks}
\end{figure}

We find that this improvement in robustness is highly correlated with the policy's Lipschitz bound. Comparing unconstrained policies to Lipschitz-bounded policies with various $\gamma$ in Figure~\ref{fig:pendulum-attacks}, there is a smooth transition from high to low robustness to sample delays and $\ell_2$-optimal adversarial attacks as the policy's Lipschitz bound increases. Interestingly, it appears that there is a ``best choice'' for $\gamma$, and that restricting it to very small values ($\gamma = 3$) harms the closed-loop performance. This is to be expected, since the optimal policy for pendulum swing-up is known to be non-smooth, hence with excessive regularization it is likely that the network's parameter space does not contain any high-performing policies. Aside from the policies with $\gamma = 3$, all models were successfully trained to approximately the same final reward (reward curves omitted for brevity). It is therefore clear from Figures~\ref{fig:pendulum-contours} to \ref{fig:pendulum-attacks} that, at least in the context of pendulum swing-up, Lipschitz-bounded policy networks significantly improve robustness to disturbances and adversarial attacks over standard, unconstrained networks.

\subsection{Comparing Architectures --- Atari Pong} \label{sec:pong-results}

\begin{figure}[!ht]
     \centering
     \begin{subfigure}[b]{0.495\textwidth}
         \centering
         \includegraphics[trim={0cm 4.3cm 0cm 0.7cm},clip,width=\textwidth]{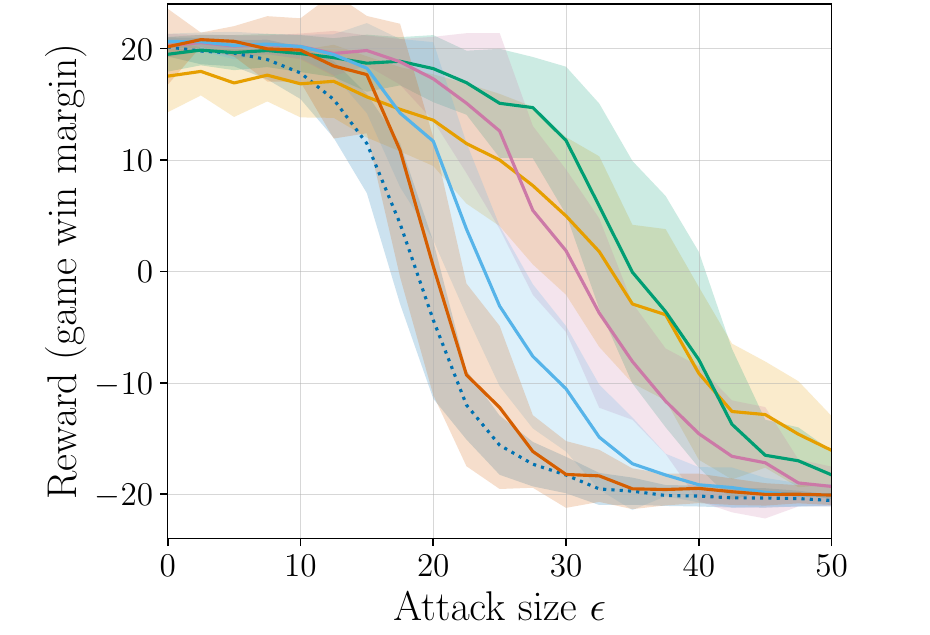}
         \caption{Uniform random noise.}
         \label{fig:pong-attacks-uniform}
     \end{subfigure}
     \begin{subfigure}[b]{0.495\textwidth}
         \centering
         \includegraphics[trim={0cm 4.3cm 0cm 0.7cm},clip,width=\textwidth]{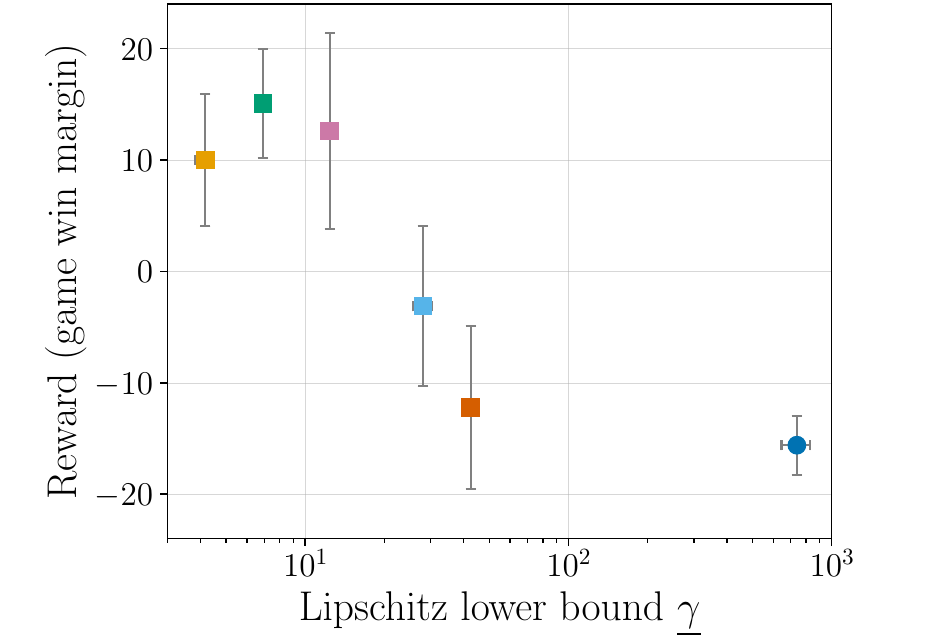}
         \caption{Cross-section at $\epsilon = 25$.}
         \label{fig:pong-attacks-uniform-lip}
     \end{subfigure}
     \begin{subfigure}[b]{0.495\textwidth}
         \centering
         \includegraphics[trim={0cm 4.3cm 0cm 0cm},clip,width=\textwidth]{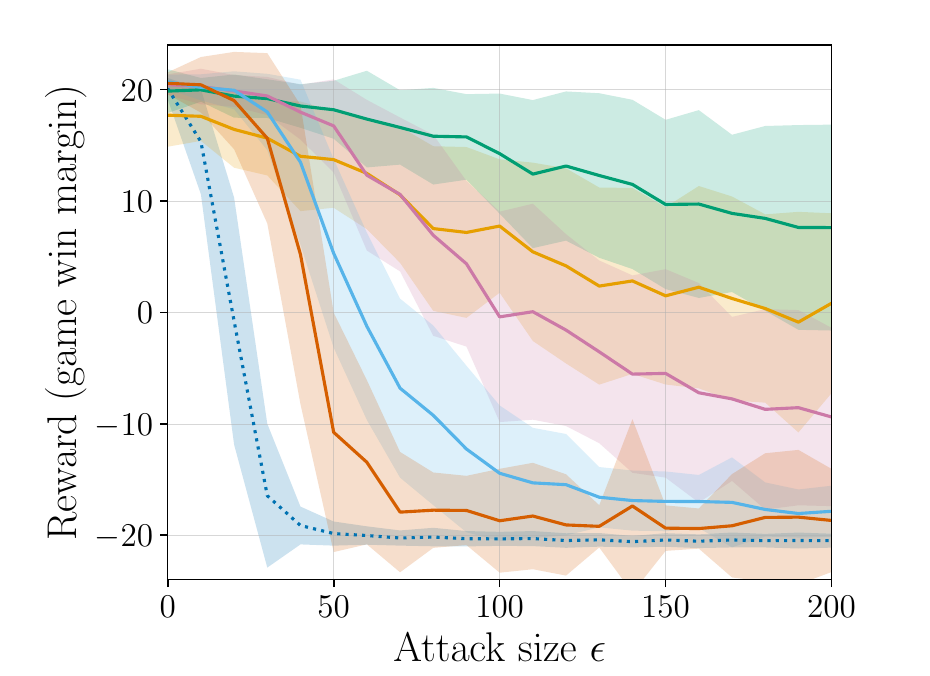}
         \caption{$\ell_2$ PGD attacks.}
         \label{fig:pong-attacks-lbdn-l2}
     \end{subfigure}
     \begin{subfigure}[b]{0.495\textwidth}
         \centering
         \includegraphics[trim={0cm 4.3cm 0cm 0cm},clip,width=\textwidth]{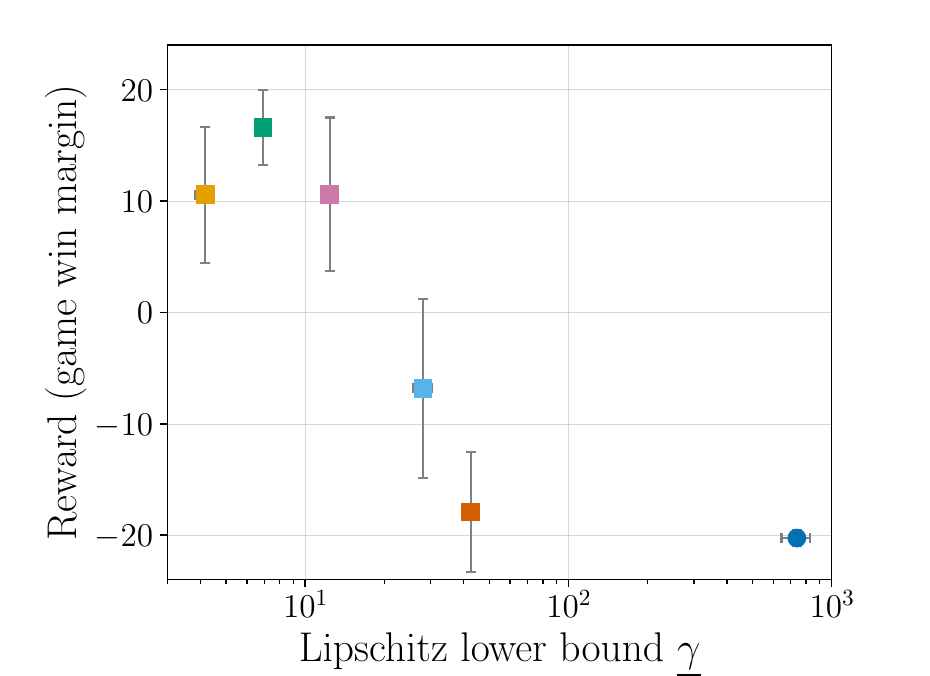}
         \caption{Cross-section at $\epsilon = 70$.}
         \label{fig:pong-attacks-lbdn-l2-lip}
     \end{subfigure}
     \begin{subfigure}[b]{0.495\textwidth}
         \centering
         \includegraphics[trim={0cm 1.5cm 0cm 0cm},clip,width=\textwidth]{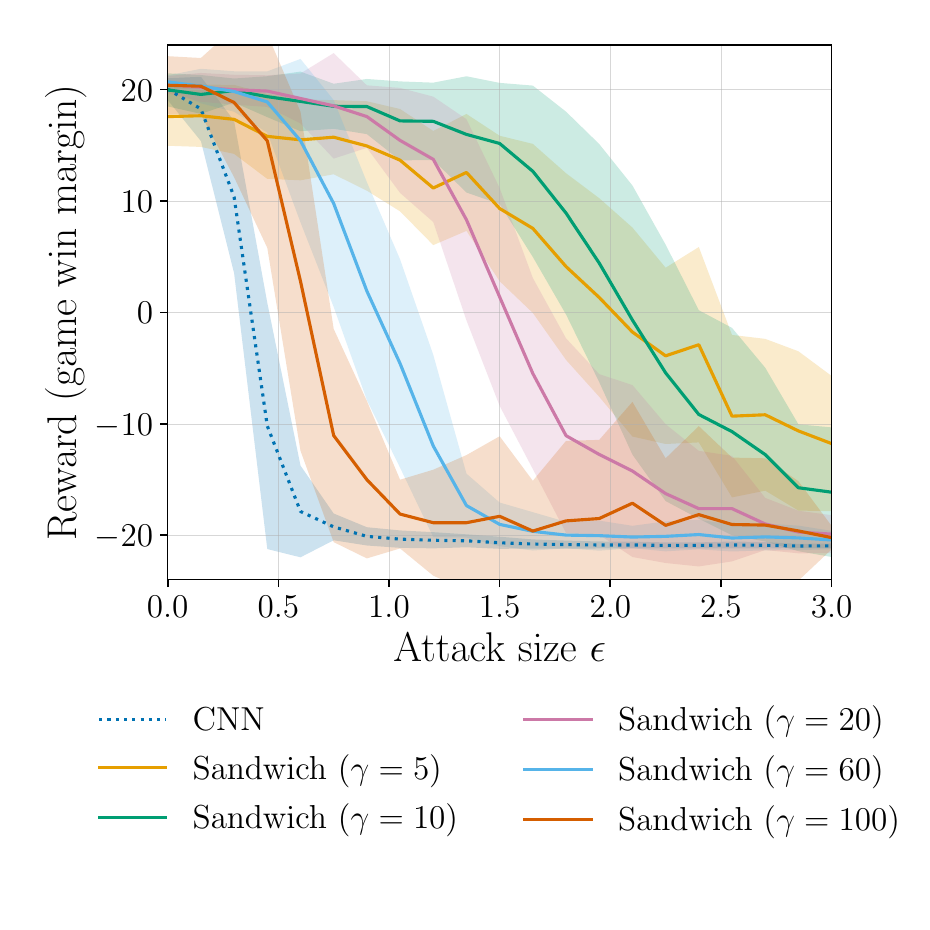}
         \caption{$\ell_\infty$ PGD attacks.}
         \label{fig:pong-attacks-lbdn-linf}
     \end{subfigure}
     \begin{subfigure}[b]{0.495\textwidth}
         \centering
         \includegraphics[trim={0cm 1.5cm 0cm 0cm},clip,width=\textwidth]{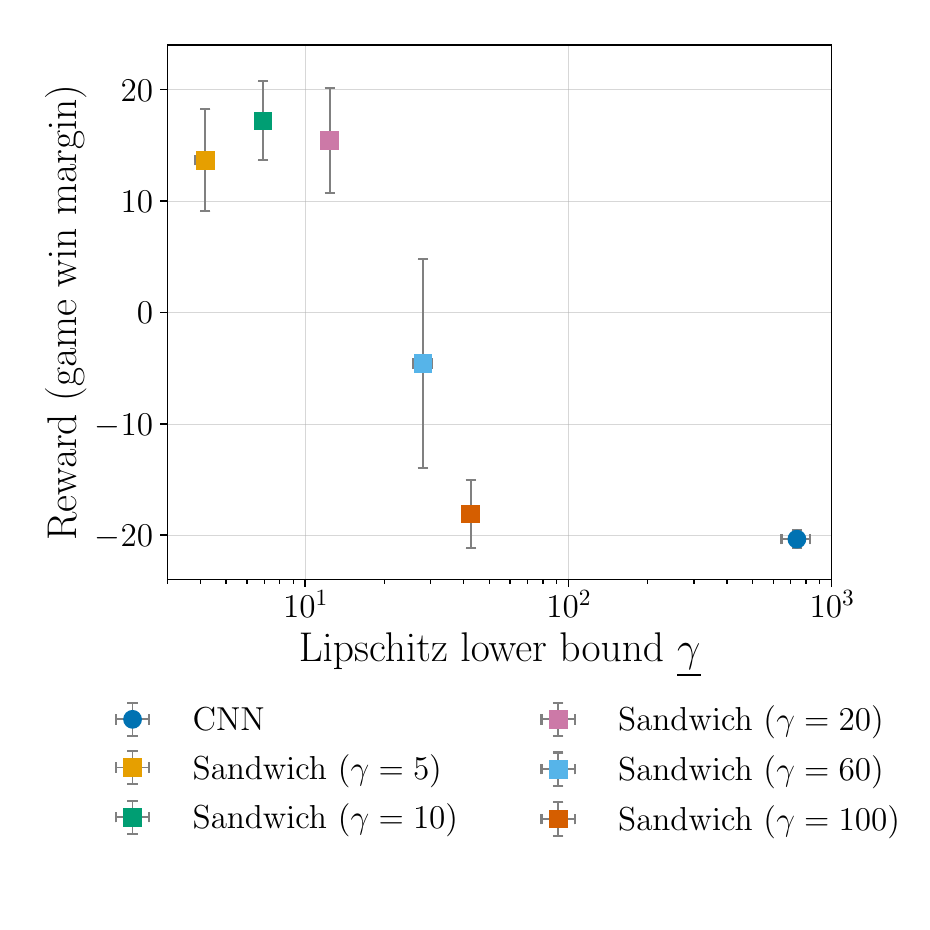}
         \caption{Cross-section at $\epsilon = 1.05$.}
         \label{fig:pong-attacks-lbdn-linf-lip}
     \end{subfigure}
     \vspace{-4mm}
    \caption{Robust performance of unconstrained (CNN) and Lipschitz-bounded (Sandwich) policies for Atari Pong. Bands and error bars show one standard deviation over 4 random model initializations.}
    \vspace{-2mm}
    \label{fig:pong-attacks}
\end{figure}

In the game of Pong, we expect small amounts of image noise to make very little difference to the state of the game and the optimal action. We would therefore hope that smooth, Lipschitz-bounded policies can improve robustness to perturbations like random noise and adversarial attacks. We see this immediately in Figure~\ref{fig:pong-attacks}, where we compare the robustness of unconstrained CNN policies and Lipschitz-bounded Sandwich policies to uniform random noise, $\ell_2$ PGD attacks, and $\ell_\infty$ PGD attacks. The same qualitative results observed for the pendulum in Figure~\ref{fig:pendulum-attacks} can be seen in Figure~\ref{fig:pong-attacks}: there is a smooth transition from high to low robustness as $\gamma$ decreases; robustness is improved not just for $\ell_2$-constrained attacks (which we expect for policies with a small $\ell_2$ Lipschitz bound), but also for random noise and $\ell_\infty$-constrained attacks; and if $\gamma$ is too small (e.g., $\gamma = 5$ here), the policy's nominal performance and robustness to perturbations is degraded. In this case, it is possible that the $\gamma = 5$ models have not finished training and could perform better if trained over more epochs (Fig.~\ref{fig:pong-reward-lbdn}).

\begin{figure}[!ht]
    \centering
    \includegraphics[trim={2cm 2cm 2cm 2cm},clip,width=\textwidth]{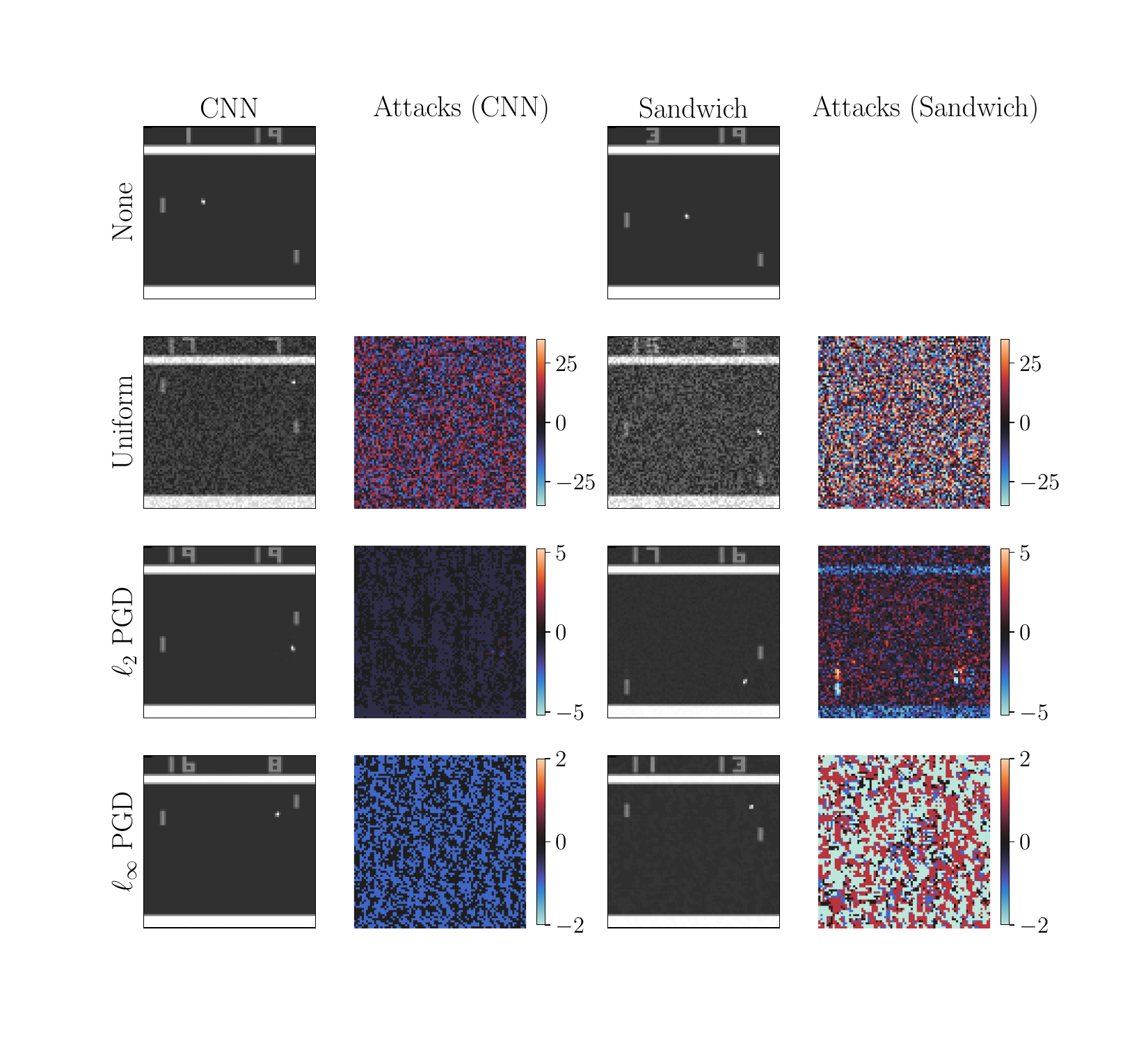}
    \caption{Adversarial examples showing the smallest attacks required to make an unconstrained (CNN) and a Lipschitz-bounded (Sandwich, $\gamma = 10$) policy lose the game. In each image, the ``computer'' controls the paddle on the left, while the policy controls the paddle on the right. All attacked frames show scenarios where the policy is about to concede a goal (puck moving to the right).} 
    \label{fig:pong-gameplay-attacks}
\end{figure}

It is interesting to look deeper into the effect of adversarial attacks on these models. Figure~\ref{fig:pong-gameplay-attacks} shows just how much of an improvement Lipschitz-bounded policy networks provide in Pong over a standard, unconstrained CNN. The CNN loses the game when subject to very small amounts of random noise and almost imperceptible adversarial attacks. In contrast, the Lipschitz-bounded policy is only beaten with a level of random noise that would even make the game difficult for a human. Moreover, highly-structured $\ell_2$-constrained attacks are required to beat the Lipschitz-bounded policy. Looking closely at Figure~\ref{fig:pong-gameplay-attacks}, successful $\ell_2$ PGD attacks try to trick the policy into thinking the opponent's paddle is shifted from where it actually is while also trying to hide the exact location of the puck. The $\ell_2$ attacks also seem to focus on the white boundary walls of the game. It is less clear why the policies should be sensitive to these features, but we hypothesize that the straight lines of the walls may appear similar to a paddle in feature space after passing through convolutional layers. There is no clear structure to the $\ell_\infty$ PGD attacks and they remain rather small, since the Lipschitz-bounded policies have a constrained $\ell_2$ Lipschitz bound, and the $\ell_2$ norm is only a loose upper bound of the $\ell_\infty$ norm in high-dimensional spaces. The additional robustness to $\ell_\infty$ PGD attacks over CNN policies is a nice bonus.

\begin{figure}[!t]
    \centering
    \begin{subfigure}[b]{0.495\textwidth}
         \centering
         \includegraphics[trim={0cm 0.5cm 0cm 0cm},clip,width=\textwidth]{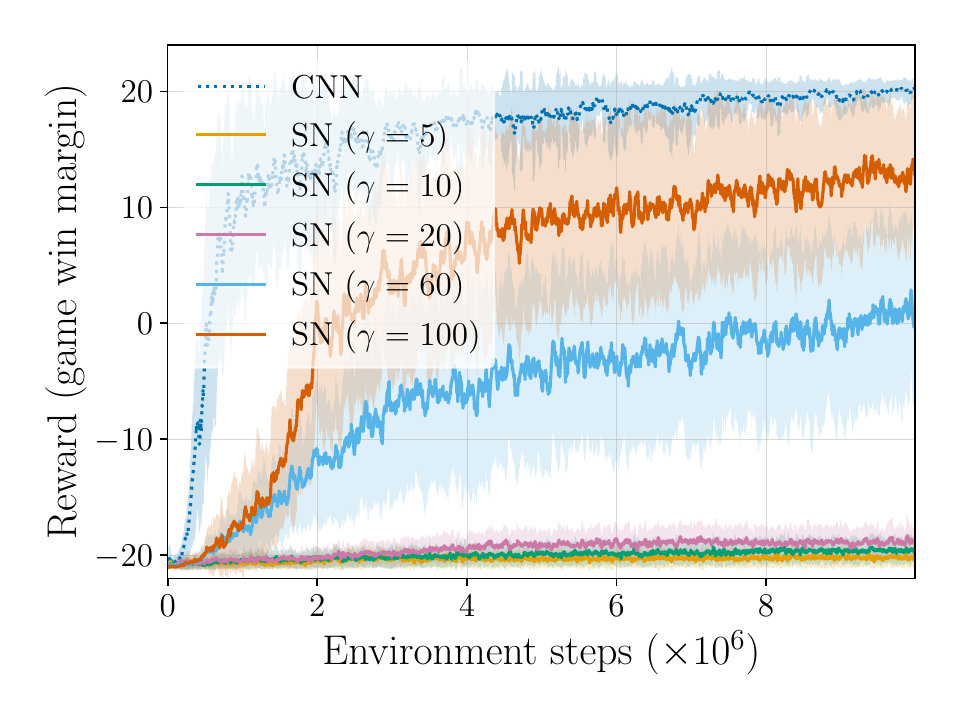}
         \caption{SN.}
         \label{fig:pong-reward-spectral}
    \end{subfigure}
    \begin{subfigure}[b]{0.495\textwidth}
         \centering
         \includegraphics[trim={0cm 0.5cm 0cm 0cm},clip,width=\textwidth]{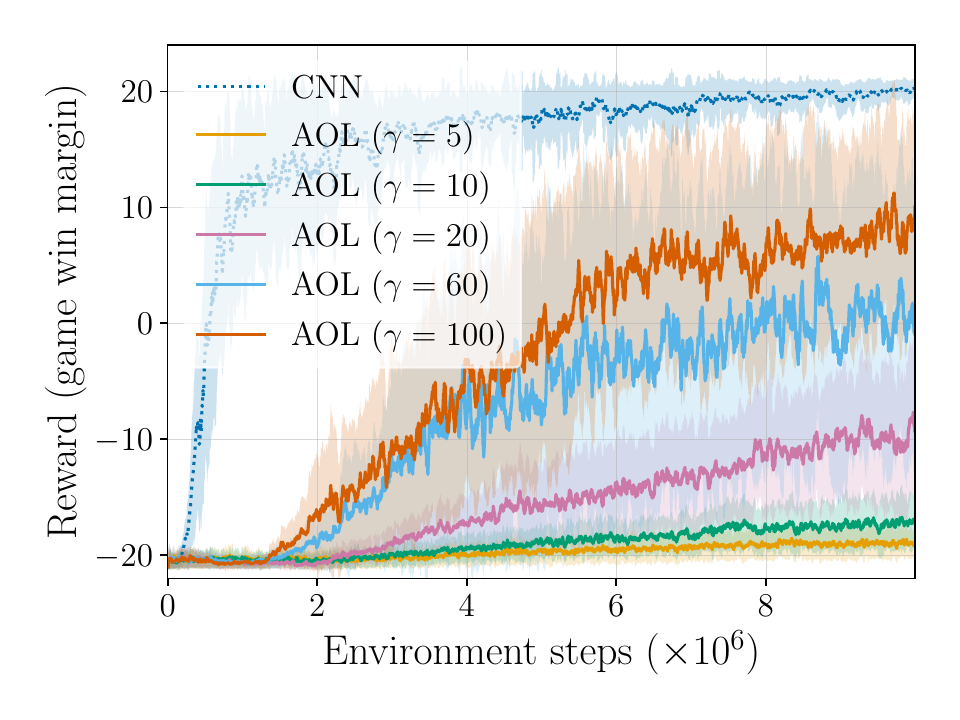}
         \caption{AOL.}
         \label{fig:pong-reward-aol}
    \end{subfigure}
    \begin{subfigure}[b]{0.495\textwidth}
         \centering
         \includegraphics[trim={0cm 0.5cm 0cm 0cm},clip,width=\textwidth]{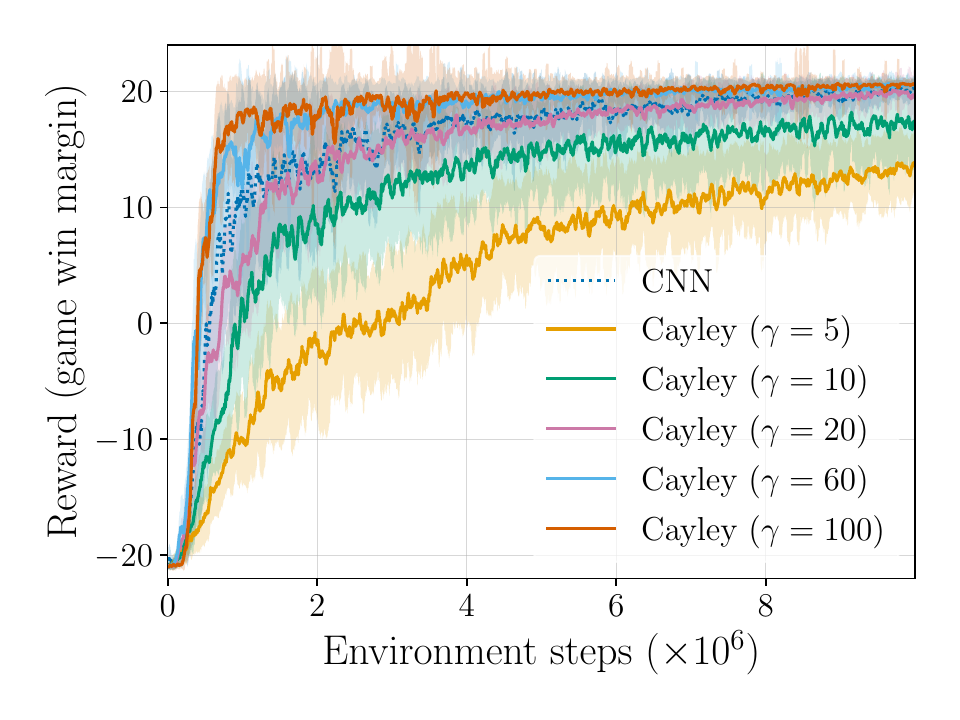}
         \caption{Cayley.}
         \label{fig:pong-reward-orthogonal}
    \end{subfigure}
    \begin{subfigure}[b]{0.495\textwidth}
         \centering
         \includegraphics[trim={0cm 0.5cm 0cm 0cm},clip,width=\textwidth]{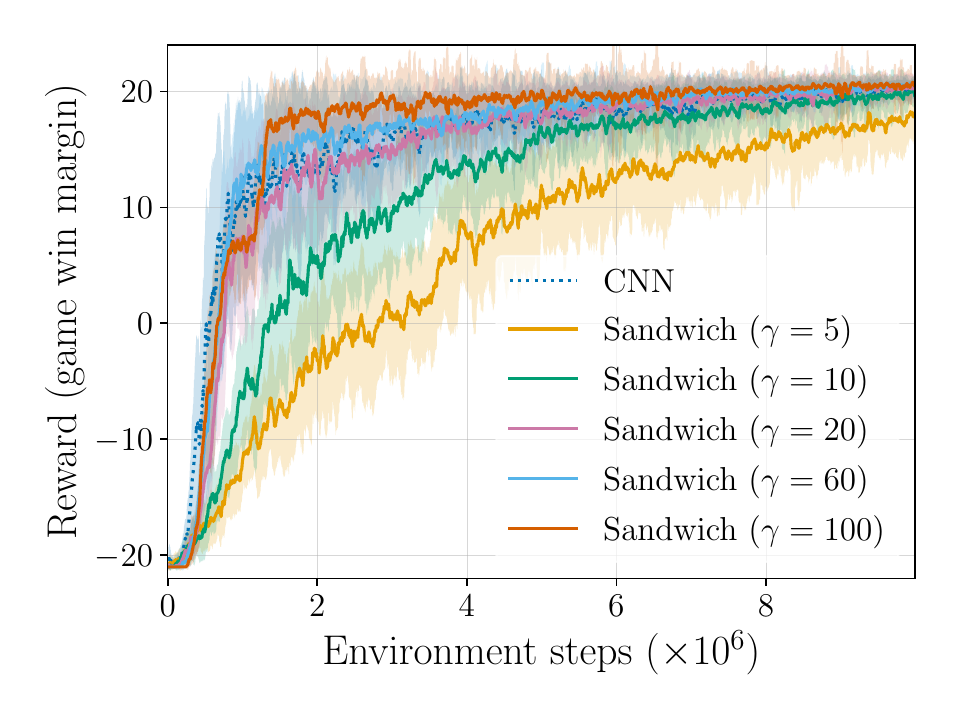}
         \caption{Sandwich.}
         \label{fig:pong-reward-lbdn}
    \end{subfigure}
    \caption{Test reward during training on Atari Pong for all layer architectures. Bands show one standard deviation over 4 random model initializations.}
    \label{fig:pong-rewards}
\end{figure}

So far, we have only investigated Lipschitz-bounded policy networks constructed from Sandwich layers to illustrate that smoother policies can improve robustness in deep RL. It turns out that in addition to choosing its upper bound $\gamma$, the layer architecture we use to bound the Lipschitz constant $\mathrm{Lip}(\kappa)$ of a policy $\kappa$ is extremely important. Figure~\ref{fig:pong-rewards} compares the training curves of all four Lipschitz-bounded policy architectures from Section~\ref{sec:background-lipnets} with unconstrained CNN policies, while Table~\ref{tab:pong-robustness} summarizes their nominal and robust performance. It is immediately clear from Figures~\ref{fig:pong-reward-spectral} and \ref{fig:pong-reward-aol} that the two layer architectures known to have conservative bounds, SN and AOL, perform poorly when $\gamma$ is small --- e.g., when $\gamma \le 20$, neither architecture ever produces a winning policy. For larger $\gamma$, the SN and AOL policies learn to win the game, but their training dynamics are extremely slow in comparison to the CNN (this is a known problem for AOL, see \cite[Sec.7]{Bernd+Lampert2022}). Moreover, Table~\ref{tab:pong-robustness} shows that the estimated lower bound $\underline\gamma$ on $\mathrm{Lip}(\kappa)$ for these policies is very small when $\gamma$ is small. This suggests that the conservative parameterization of SN and AOL layers restricts their parameter spaces to a small set of very smooth models which does not include high-performing policies. In contrast, Figures~\ref{fig:pong-reward-orthogonal} and \ref{fig:pong-reward-lbdn} show that the two layer architectures with much tighter bounds on $\mathrm{Lip}(\kappa)$, Cayley and Sandwich, perform quite well even for small choices of $\gamma$. The choice of $\gamma$ still seems to have a strong impact on the training dynamics for these layers --- as $\gamma$ increases, so too does the speed at which the policies converge on a winning strategy. This raises interesting questions about the coupling between a policy's Lipschitz constant and the exploration of its parameter space, which we leave for future work.

\begin{table}[!t]
    \centering
    \begin{tabularx}{\linewidth}{
    >{\raggedright\arraybackslash}X | 
    >{\hsize=.5\hsize\centering\arraybackslash}X >{\hsize=.5\hsize\centering\arraybackslash}X |
    >{\centering\arraybackslash}X |
    >{\centering\arraybackslash}X >{\centering\arraybackslash}X >{\centering\arraybackslash}X
    }
        \toprule 
        \textbf{Policy} & \multicolumn{2}{c|}{\textbf{Lipschitz}} & \textbf{Reward} & \multicolumn{3}{c}{\textbf{Smallest winning attack size $\epsilon$ ($\uparrow$)}} \\
        \midrule
         & \textbf{$\gamma$} & \textbf{$\underline\gamma$} & Unperturbed & Uniform & $\ell_2$ PGD & $\ell_\infty$ PGD \\
        \midrule
        CNN & - & 738 & 20.1 & 18.7 & 19.5 & 0.38 \\
        \midrule
        \multirow{5}{*}{SN} & 5 & 1.14 & -20.2 & - & - & - \\
        & 10 & 2.78 & -19.4 & - & - & - \\
        & 20 & 6.05 & -18.7 & - & - & - \\
        & 60 & 24.7 & 1.21 & 15.0 & 9.80 & 0.21 \\
        & 100 & 42.4 & 14.8 & 21.5 & 21.3 & 0.49 \\
        \midrule
        \multirow{5}{*}{AOL} & 5 & 1.40 & -18.9 & - & - & - \\
        & 10 & 2.95 & -16.9 & - & - & - \\
        & 20 & 7.62 & -9.28 & - & - & - \\
        & 60 & 15.3 & 1.31 & 12.4 & 19.2 & 0.52 \\
        & 100 & 17.4 & 8.85 & 23.6$^*$ & 42.3 & 1.20$^*$ \\
        \midrule
        \multirow{5}{*}{Cayley} & 5 & 4.56 & 13.6 & 23.7 & 106 & 1.52 \\
        & 10 & 8.99 & 17.8 & 28.7 & 154 & 1.59 \\
        & 20 & 16.6 & 20.0 & 23.1 & 129$^*$ & 1.36 \\
        & 60 & 34.8 & \textbf{20.6}$^*$ & 20.9 & 67.4$^*$ & 0.98$^*$ \\
        & 100 & 47.0 & 20.4$^*$ & 17.1 & 54.0$^*$ & 0.76 \\
        \midrule
        \multirow{5}{*}{Sandwich } & 5 & 4.16 & 17.5$^*$ & 33.4$^*$ & 183$^*$ & 2.01$^*$ \\
        & 10 & 6.90 & 19.5$^*$ & \textbf{35.0}$^*$ & $\mathbf{>200}$$^*$ & \textbf{2.08}$^*$ \\
        & 20 & 12.4 & 20.2$^*$ & 30.8$^*$ & 99.2 & 1.53$^*$ \\
        & 60 & 28.0 & \textbf{20.6}$^*$ & 23.9$^*$ & 58.1 & 0.94 \\
        & 100 & 42.5 & 20.2 & 20.1 & 43.3 & 0.63 \\
        \bottomrule
    \end{tabularx}
    \vspace{3mm}
    \caption{Averaged results (4 random model initialisations and 20 random game and attack seeds) on performance and robustness of policy networks trained on Atari Pong. $\gamma$ is the certified Lipschitz upper-bound for a network and $\underline{\gamma}$ is its empirically-estimated lower-bound. The unperturbed reward is the final mean test reward achieved during training. Results for each attack strategy are the smallest average attack size $\epsilon$ required to beat the policy (i.e., reward $<0$). Attack results are not provided for policies that did not learn a positive reward.
    \textbf{Bold} values indicate the overall best performing models in each column. Values with an * indicate the best performing models for each $\gamma$.}
    \label{tab:pong-robustness}
    \vspace{-4mm}
\end{table}

Looking closer at Table~\ref{tab:pong-robustness} reveals that simply having a tight Lipschitz bound (i.e., close lower and upper bounds $\underline\gamma$ and $\gamma$) is not the only factor contributing to a policy's performance and robustness. Instead, there appears to be an advantage to using ``expressive'' policy networks, particularly when high robustness (small $\gamma$) is required. Table~\ref{tab:pong-robustness} indicates that both the Sandwich and Cayley policies exhibit strong performance and robustness for $\gamma \ge 20$. Either layer architecture is therefore a suitable choice for reasonable improvement over existing methods in robust RL like SN. For smaller $\gamma<20$, however, Sandwich policies are far superior, and with $\gamma = 10$ they are the most robust of any policy across all three input perturbations while still achieving a strong unperturbed reward of 19.6. This is despite the fact that Cayley policies often have a tighter Lipschitz bound than the Sandwich policies (take $\gamma = 10$ as an example). We suggest that this is due to the less conservative parameterization of the Sandwich layers. Each of the SN, AOL, and Cayley policies are composed of linear layers with a spectral norm of approximately 1 (exactly 1 for Cayley). In contrast, Sandwich layers have no direct restriction on their spectral norm \cite[Fig.~4]{Wang+Manchester2023}, and instead constrain $\mathrm{Lip}(\kappa)$ via a composition of nonlinear layers which are a complete parameterization of \textit{all} networks satisfying the tightest known bounds on the Lipschitz constant of DNNs \cite{Fazlyab++2019}, of which 1-Lipschitz linear layers are a special case. This expressivity allows the Sandwich models to converge on policy networks that are both performant and robust even when their parameter space is restricted by a small value of $\gamma$, allowing finer control over the performance-robustness trade-off in deep RL.

%
%
\section{Conclusions} \label{sec:conc}

This paper has studied the robustness benefits of Lipschitz-bounded policy networks in deep RL. We have found that policy networks with small Lipschitz bounds are significantly more robust to perturbations such as disturbances, random noise, and targeted adversarial attacks. Moreover, we have observed that choosing a policy with non-conservative Lipschitz bounds and an expressive layer architecture gives the user finer control over the performance-robustness trade-off than existing methods based on spectral normalization. This raises interesting questions for future study, such as whether Lipschitz-bounded policy networks can augment or alleviate the need for adversarial training, and whether the observed benefits can be transferred to real-world robotic systems.

%
%

\begin{credits}
\section*{Acknowledgments}
This work was supported in part by Google LLC, the Australian Research Council (DP230101014), and the Centre for Advanced Defence Research in Robotics and Autonomous Systems (CADR-RAS). The authors would also like to thank Jack Naylor for his assistance with computational resources.
\end{credits}

\bibliographystyle{splncs04.bst}
\bibliography{references}

\ifdefined\Arxiv
\newpage
\clearpage
\appendix
\section{Training details} \label{app:training-details}

\subsection{Pendulum swing-up experiments} \label{app:pendulum}

The pendulum was modeled in MuJoCo XLA (MJX), a fully-differentiable implementation of the MuJoCo physics simulator \cite{Todorov++2012} written in JAX \cite{jax2018github} that allows users to take gradients through the entire closed-loop system. We trained unconstrained MLP and Lipschitz-bounded (Sandwich) policies using the PPO implementation in \cite{Freeman++2021}, which leverages the scalability of JAX to massively multi-thread parallel physics simulations on a single workstation GPU. Hyperparameters were tuned by varying each parameter one at a time and choosing the best-performing parameters for an unconstrained MLP. The same hyperparameters were used to train Lipschitz-bounded policies without any further tuning. Our chosen hyperparameters can be found on our GitHub repository\footnotemark{}. We trained 10 policies for each model architecture and choice of $\gamma$, each with a different random seed for model initialization. Results showing robust performance to sample delays and adversarial attacks in Figure~\ref{fig:pendulum-attacks} were averaged over the 10 policies and 1024 pendulum environments starting from random initial states.
MLP networks were composed of 4 linear layers of 32 hidden nodes. Lipschitz-bounded policies were composed of 4 Sandwich layers of 21 hidden nodes to ensure the two model architectures had a similar number of trainable parameters. We chose tanh activations for all policies in accordance with \cite{Andrychowicz++2021}.

\footnotetext{\scriptsize{\url{https://github.com/nic-barbara/Lipschitz-RL-MJX/blob/main/scripts/pendulum/train_2_train_models.py}}}

\subsection{Atari Pong experiments} \label{app:pong}

We trained unconstrained CNN models and the Lipschitz-bounded policy architectures (SN, AOL, Sandwich, Cayley) across 4 random model initialisations using the PPO implementation in \cite{huang2022cleanrl} and the ALE/Pong-v5 environment from \cite{weng2022envpool}. We used the default hyperparameters chosen for CNN policies in \cite{huang2022cleanrl} for all policy architectures. We could not use the reward gradient to directly optimize \eqref{eqn:optimal-attack} as the Pong environment is not differentiable, hence we implemented attacks with the PGD method \cite{ding2019advertorch}. Robust performance to noise and adversarial attacks in Figure~\ref{fig:pong-attacks} and Table~\ref{tab:pong-robustness} were averaged over the 4 policies and 20 games of Pong, each with a different random seed. Further details on the network architecture for each layer type can be found on our GitHub repository\footnotemark{}. 

\footnotetext{\scriptsize{\url{https://github.com/nic-barbara/Lipschitz-RL-Atari/blob/main/liprl/atari_agent.py}}}
\fi

\end{document}